\documentclass[runningheads]{llncs}
\usepackage[T1]{fontenc}
\usepackage{graphicx}
\usepackage{color}
\usepackage{comment}
\usepackage{amsmath,amsfonts,amssymb}
\usepackage{enumitem}
\usepackage{shuffle}
\usepackage{stmaryrd}
\usepackage{mathtools}
\usepackage{hyperref}

\usepackage{algorithm}
\usepackage{algpseudocode}
\usepackage{caption}
\usepackage{subcaption} 

\begin{document}
\title{Hypergraphs on high dimensional time series sets using signature transform.}
%
%
%
\author{Rémi Vaucher\inst{1,2}\orcidID{0009-0005-4315-7870}\\
\and Paul Minchella\inst{3,4}}
\authorrunning{R. Vaucher et al.}
%
\institute{Laboratoire ERIC, Bron 69676, France\\
\and
Halias Technologies, Meylan 38240, France, \email{remi.vaucher@halias.fr}\\
\and Université Claude Bernard Lyon 1, 69100 Villeurbanne, France \email{paul.minchella@lyon.unicancer.fr}\\
\and Centre Léon Bérard, Lyon 69008, France}
\maketitle              
\begin{abstract}
In recent decades, hypergraphs and their analysis through Topological Data Analysis (TDA) have emerged as powerful tools for understanding complex data structures. Various methods have been developed to construct hypergraphs—referred to as simplicial complexes in the TDA framework—over datasets, enabling the formation of edges between more than two vertices.

This paper addresses the challenge of constructing hypergraphs from collections of multivariate time series. While prior work has focused on the case of a single multivariate time series, we extend this framework to handle collections of such time series. Our approach generalizes the method proposed in \cite{chretien2024leveraging} by leveraging the properties of signature transforms to introduce controlled randomness, thereby enhancing the robustness of the construction process. We validate our method on synthetic datasets and present promising results.

\end{abstract}
%
%

\section{Introduction: Computing hypergraph on a multivariate time series using signature transform.}

\subsection{Motivation and Background}
The construction of hypergraphs from time series data has gained significant attention in recent years due to their ability to capture higher-order relationships that traditional pairwise graphs cannot represent. While existing methods have addressed the construction of hypergraphs from single multivariate time series, the extension to multiple multivariate time series presents unique challenges and opportunities.

For instance, in epidemiology, multiple time series may represent symptom dynamics across different geographic regions. Understanding the higher-order interactions between these symptom patterns can provide insights into disease propagation and emergence of novel pathologies. Similarly, in financial markets, analyzing multiple asset price series simultaneously can reveal complex market dependencies that simple correlation analysis might miss.

\subsection{Related Work.}

Several approaches have been proposed for constructing hypergraphs from time series data \cite{petri2014homological,santoro2023higher,stolz2017persistent,stolz2021topological}. However, Algorithm \ref{infcompsimp} appears to be the only approach capable of being extended to a new type of task: construct a hypergraph on top of $\{Y_1,...,Y_N\}$ where \textbf{each $Y_i$ is itself multivariate.}

\subsection{The signature transform.}

Consider $X:[a,b]\rightarrow\mathbb{R}^d$ a path \textbf{of bounded variation}:
\begin{align*}
    X & \in BV([a,b],\mathbb{R}^d) \\
    & =\left\{X:[a,b]\rightarrow \mathbb{R}^d\in L^1,\quad\sup_{P\in \mathcal{P}}\sum\limits_{i=0}^{n_P-1}\|X(t_{i+1})-X(t_i)\|<\infty\right\}
\end{align*}
where $\mathcal{P}$ is the set of all subdivisions of $[a,b]$.
The \textbf{signature of $X$}, noted $S_{[a,b]}(X)$, is the tensor sequence of increasing dimension
\begin{align*}
    S_{[a,b]}(X)=1\oplus S^{(1)}(X)\oplus S^{(2)}(X)\oplus \dots \oplus S^{(k)}(X)\oplus\dots\in T(\mathbb{R}^d)
\end{align*}
with $S_{[a,b]}^{(k)}(X)\in(\mathbb{R}^d)^{\otimes k}$ and
\begin{align*}
    \left(S^{(k)}(X)\right)_{i_1\dots i_k}=\idotsint\limits_{a<t_1<\dots<t_k<b}dX^{i_1}(t_{i_1})\dots dX^{i_k}(t_{i_k}) 
\end{align*}
Finally, we need one more definition:

\begin{definition}
    Let $X\in BV([a,b],\mathbb{R}^d)$ a bounded variation path. An \textbf{augmentation of the paths $X$} is an application
    \begin{align*}
        \pi_k:BV([a,b],\mathbb{R}^d)\rightarrow BV([a,b],\mathbb{R}^{D})
    \end{align*}
    with $d<D$.
\end{definition}
Augmentation allows us to embed lower-dimensional paths into higher-dimensional spaces while preserving essential geometric properties.
One of the most known augmentation is the \textbf{lead-lag augmentation} \cite{fermanian2021learning,lyons2022signature}:
\begin{align*}
    LL:BV([a,b],\mathbb{R}^d)\rightarrow BV([a,b],\mathbb{R}^{2d})
\end{align*}
Signature has many useful properties, and we invite the reader to discover everything about signature in \cite{chevyrev2016primer,fermanian2021learning,lyons2022signature}.

\subsection{Hypergraph on multivariate time series.}

In \cite{chretien2024leveraging}, the authors use the Algorithm \ref{infcompsimp} to construct a \textbf{hypergraph} over a multivariate time series $X = (X^1,...,X^d)$.

\vspace{.3cm}

\textbf{Remark:} In the following sections, we adopt the topological terminology and refer to hypergraphs as \textbf{simplicial complexes} to align with the TDA framework. The final objective is to use TDA framework, like filtration, on it. This problem will be addressed in future work.

\vspace{.3cm}

Consider a set of $n$ vertices $V=\{v_1,\dots,v_n\}$.
\begin{definition}
    For $k<n$, a \textbf{$k$-simplex} $\sigma_k$ of $V$ is the collection of a subset $V_k$ of length $k+1$ and all its subsets. The \textbf{geometric realization} of a $k$-simplex is the convex hull $C$ of $k+1$ points, such that $\dim(C)=k$. A \textit{face} (of dimension $l$) in $\sigma_k$ is a collection of set in $\sigma_k$ that form a $l$-simplex ($l\leq k)$.
\end{definition}

\textbf{Notation:} In the following, for convenience, we will often note $[v_{i_1}\dots v_{i_k}]$ the simplex over the vertices $\{v_{i_1},\dots,v_{i_k}\}$ following the notations of \textbf{hyperedges}.

\begin{definition}
    A \textbf{simplicial complex} $\mathcal C$ on $V$ is a collection of simplexes (of $V$) such that for every $\sigma^i\in\mathcal C$, there exists $j$ with $\sigma^i\cap\sigma^j$ a sub-simplex of both $\sigma^i$ and $\sigma^j$.
\end{definition}

To build such a complex from a multivariate time series, \cite{chretien2024leveraging} consider $X\in BV([a,b],\mathbb{R}^d)$ as an element of $\bigg (BV([a,b],\mathbb{R})\bigg)^d$. 

Then, the considered set of vertices is $\{X^1,...,X^d\}\subset BV([a,b],\mathbb R)$. The algorithm \ref{klink} uses the notion of $k$-dimensional Link:

\begin{definition}
\label{link}
    Consider $k \in N^*$ and $\sigma$ a $k$-simplex in a simplicial complex $\mathcal C$. Its \textbf{link} $\text{Lk}(\sigma,\mathcal C )$ in $\mathcal C$ is the set of all faces $\tau\subset\mathcal C$ such that:
    \begin{itemize}
        \item $\sigma\cap \tau = \varnothing$
        \item $\sigma\cup\tau$ is a face of $\mathcal{C}$
    \end{itemize}
\end{definition} and the $k$-dimensional Link is just the subset of all $k$-dimensional elements of $\text{Lk}(\sigma,\mathcal{C})$.

\begin{algorithm}[!h]
\begin{algorithmic}
    \Require A set of vertices $V=\{v_1,...,v_N\}$; an index $i$; a dimension $k$; an interval $[a,b]\subset [0;T]$.
    \Ensure{The $k$-dimensional link of $v_i$ in $\mathcal{C}$.}
    \State $\bullet$ Computation of signature $S_i$ for vertex $v_i$ over $[a,b]$ (eventually \textbf{augmented} for dimension coherence).\\
    \State $\bullet$ For every subset $v_{i_1},...,v_{i_k}$ with length $k$, compute the signature $S_{i_1\dots i_k}$ of the $k$-dimensional path defined by $\{v_{i_1},\dots,v_{i_k}\}$ over $[a,b]$.\\
    \State $\bullet$ Thanks to a variable selection algorithm, compute the regression coefficients for 
    \begin{align}
    S_i = \sum\limits_{\{v_{i_1},...,v_{i_k}\}\subset V\setminus\{v_i\}}\beta_{i_1\dots i_k}S_{i_1\dots i_k}.
    \label{lasso}
    \end{align}
  \For{$\{v_{i_1},...,v_{i_k}\}\subset V\setminus\{v_i\}$}
    \State \If{$\beta_{i_1\dots i_k}\neq 0$ \textbf{and} $R^2>0.67$ (for a model quality insurance)}
    \State Include the $(k-1)$-simplex $[v_{i_1}\dots v_{i_k}]$ in the $(k-1)$-dimensional link of $v_i$.
    \EndIf
  \EndFor
  \caption{Prediction of the $k$-dimensional Link of $v_i$ in $\mathcal{C}$.}
  \label{klink}
\end{algorithmic}
\end{algorithm}

\begin{algorithm}[!h]
\begin{algorithmic}
    \Require A set of vertices $V=\{v_1,\dots ,v_k\}$; a dimension $K$; an interval $[a,b]\subset[0;T]$.
    \Ensure The explainability simplicial complex $\mathcal{C}$ over $V$.
    \For{$v_i\in V$}
    \For{$k\leq K-1$}
    \State Predict the $k$-dimensional Link of $v_i$ in $\mathcal{C}$ thanks to algorithm \ref{klink}
    \EndFor
    \For{$\sigma=[v_{i_1}\dots v_{i_k}]$ in the $k$-dimensional Link for $v_i$}
    Include the simplex $[v_iv_{i_1}\dots v_{i_k}]$ in $\mathcal{C}$.
    \EndFor
    \EndFor
    \caption{Prediction of the explainability simplicial complex over $V$.}
    \label{infcompsimp}
\end{algorithmic}
\end{algorithm}

\section{Modifications of the algorithm.}

\subsection{Fitting the new data format.}
Notably, no fundamental modifications are required: each vertex in Algorithm \ref{klink} is now a multidimensional path, which the signature transform can still accommodate. The primary challenge that may arise concerns \textbf{dimensional consistency}. 

\begin{itemize}
    \item \textbf{Case 1:} When $Y_i\in BV([a,b],\mathbb{R}^d)$ for every $i\in\{1,...,n\}$, the dimensions are the same, and no modifications are needed.

    \item \textbf{Case 2:} In the case of \textbf{missing dimensions}, \textit{i.e.}, each $Y_i$ are supposed to have the same joint distribution but at least one dimension is missing for at least one $Y_i$, we propose two approaches:
    \begin{enumerate}
        \item \textbf{Projection approach:} if $Y_i\in BV([a,b],\mathbb{R}^d)$ for at most $n-1$ indices $i\in\{1,...,n\}$, and there is at least one $j\in\{1,...,n\}$ such that $Y_j\in BV([a,b],\mathbb{R}^{d_j})$, with $d_j<d$, it is possible to constrain the study a projection of each $Y_i$ on the exact dimensions of $Y_j$. 
    
    Mathematically speaking, if $Y_j = (Y_j^1,...,Y_j^{k-1},Y_j^{k+1},...,Y_j^{d})$, and defining 
    \begin{align*}
    \begin{array}{rcl}
    \pi_k:BV([a,b],\mathbb{R}^d)&\rightarrow& BV([a,b],\mathbb{R}^{d-1})\\ Y&\mapsto&(Y^1,...,Y^{k-1},Y^{k+1},...,Y^d)
    \end{array}
    \end{align*}
    we build our complex on the set 
    \begin{align}
        \bigg\{\pi_k(Y_1),...,\pi_k(Y_{j-1}),Y_j,\pi_k(Y_{j+1}),...,\pi_k(Y_n)\bigg\}.
    \end{align}
    \item \textbf{Augmentation approach:} 
    Another method consists of augmenting $Y_j$: considering an application 
    \begin{align*}
    \text{Aug}_d:BV([a,b],\mathbb{R}^{d_j})\rightarrow BV([a,b],\mathbb{R}^d)
    \end{align*}
    we use $\tilde{Y_j}$ instead of $Y_j$. In this case, we build our complex on the set 
    \begin{align}
    \bigg\{Y_1,...,Y_{j-1},\tilde{Y_j},Y_{j+1},...,Y_n\bigg\}.     
    \end{align}
    This method can be effective in the case of $Y_i$ being solution of a controlled differential equation: if $Y_i^k$ could be explicative of one $Y_i$'s channel, we do not want to delete $Y_i^k$.
    \end{enumerate}

    \item \textbf{Case 3:} In the case of \textbf{non-homogeneous dimensions}, \textit{i.e.},\\ $Y_i\in BV([a,b],\mathbb{R}^{d_i})$ for all $i\in\{1,...,n\}$, it is not feasible to reduce dimensionality arbitrarily, as the recorded physical quantities differ fundamentally. Therefore, we need to use an augmentation to create homogeneity.
\end{itemize}

\subsection{Bringing randomness in the algorithm.}

To enhance the robustness of our methodology, we incorporate controlled randomness through subsampling of time points. This approach addresses the sensitivity of the signature transform to noise and provides a probabilistic framework for hypergraph construction.

Let $\Sigma=\{a=t_0,t_1,...,t_{N-1},t_N=b\}$ be the sampling times of $Y_i$, for all $Y_i$ (potentially, each $Y_i$ can have its proper discretization of $[a,b]$, as long as every $Y_i$ are defined on $[a,b]$). For a given subset $\sigma\in\Sigma$, we predict the hyperadjacency tensors:

\begin{definition}[Hyper-adjacency tensors.]
    Let $\mathcal{C}$ be a simplicial complex whose vertices are $\{v_1,...,v_n\}$. The \textbf{hyper-adjacency tensors} are a sequence of tensors $\left( A_i\right)_{2\leq i\leq n}$ with 
    \begin{align*}
    A_i & \in\mathbb{R}^{\underbrace{n\times n\times \dots \times n}_{i\text{ times}}}    
    \end{align*} 
    such that
    \begin{align*}
        (A_i)_{j_1,...,j_i}=\begin{cases}
            1 & \text{if } [v_{j_1}\dots v_{j_i}]\in \mathcal{C}\\
            \\
            0 & \text{otherwise}
        \end{cases}
    \end{align*}
\end{definition}

\textbf{Remark:} $A_2$ is just the well known adjacency matrix of the skeleton of $\mathcal{C}$ (containing only 1-dimensional simplex, \textit{i.e.}, edges $[v_iv_j],\quad i\neq j$).

\vspace{.3cm}

So, for a given sample times subset $\sigma$, we build $(A_{\sigma,i})_{2\leq i\leq n}$. We hypothesize that averaging the hyper-adjacency tensors over randomly selected subsets of time points yields a robust estimate of the probability of each potential hyperedge, that we can summarize in a sequence of tensor $(\mu_{A_i})_{2\leq i\leq n}$ that we called \textbf{probability tensors}. The algorithm becomes:

\begin{algorithm}
\begin{algorithmic}
    \Require A set of vertices $V=\{v_1,\dots ,v_n\}$; a dimension $K$; a discretization $\Sigma=\{t_0,\dots,t_N\}$; a number of tries $N_{tries}$; a size for subsets $l$.
    \Ensure The probability tensors of the simplicial complex over $V$.
    \For{$j \in \{1,\dots, N_{tries}\}$:}
    \State $\bullet$ Select $l$ timesteps $\sigma_j=\{t_{j_1},\dots, t_{j_l}\}\subset\Sigma$ under uniform distribution.
    \State $\bullet$ Interpolate or restrict each $Y_i$ to the selected timesteps $\sigma_j$.
    \State $\bullet$ Compute $V_{\sigma_j}$ the set of vertices restricted to $\sigma_j$.
    \State $\bullet$ Predict the simplicial complex $\mathcal{C}_{\sigma_j}$ over $V_{\sigma_j}$ thanks to algorithm \ref{infcompsimp}.
    \State $\bullet$ Compute the sequence of hyperadjacency tensors $\left(A_{i,\sigma_j}\right)_{2\leq i\leq K}$
    \EndFor
    \State $\bullet$ Compute the sequence of probability tensors as $\bigg(\frac{1}{N_{tries}}\sum\limits_{j=1}^{N_{tries}}A_{i,\sigma_j}\bigg)_{2\leq i\leq K}$
    \caption{Prediction of the probability tensors.}
    \label{probtensors}
\end{algorithmic}
\end{algorithm}
These probability tensors can be used to:
\begin{itemize}
    \item To create a fixed simplicial complex using thresholding methods, 
    \item To predict explainability simplicial complex over $V$ distribution using a Metropolis Hastings algorithm. 
\end{itemize}

In the next section, we test our methods using a thresholding selection.

\section{Experiments}
\label{experiments}

We evaluate our algorithm using synthetic data. To fit our framework, we use the following approach\footnote{The implementation is available at \url{https://github.com/RemiVaucher/Thesis} section "Hypergraph test"}:

\begin{itemize}
    \item We generate $\{Y_1,...,Y_n\}$ with $Y_i\in BV([0;1],\mathbb{R}^2)$, sampled every $\Delta t=\frac{1}{100}$. 
    \item We restrict our experiments to 1-dimensional simplicial complexes (higher dimension will be addressed in future works).
    \item For all $i\in \{2,\dots,n-1\}$, $Y_i$ is the solution of the following ODE:
    \begin{align}
        dY_i(t)=B_1Y_i(t-h)+B_2Y_{i-1}(t-h)-B_2Y_{i+1}(t-h)+\epsilon
        \label{equadiff}
    \end{align}
    with $B_1=\begin{pmatrix}
        1&c\\
        c& 1
    \end{pmatrix}$, $B_2=\begin{pmatrix}
        c & 0\\
        0 & c
    \end{pmatrix}$, $c\in \mathbb{R}^*$, and $\epsilon\sim \mathcal{N}(0,\sigma^2)$.
    We fixed\\ $Y_1(0)\sim \mathcal{N}(0,\sigma_{start}^2Id)$ and $Y_i(0)=(0,0)$ for all $i\neq 1$.
    In such configuration, the expected hyper-adjacency tensor $A_2$ is
    \begin{align*}
        A_2=\text{Id}_{n\times n}+D_{n\times n,-1}+D_{n\times n,1}
    \end{align*}
    with $D_{n\times n,1}=D_{n\times n,-1}^T=\begin{pmatrix}
        0 & 1 & 0 & \dots & 0 & 0\\
        0 & 0 & 1 & \dots & 0 & 0\\
        \vdots & & \dots & & & \vdots\\
        0 & 0 & 0 & \dots & 0 & 1\\
        0 & 0 & 0 & \dots & 0 & 0
    \end{pmatrix}$
    \begin{figure}[!h]
        \centering
        \includegraphics[width=0.65\linewidth]{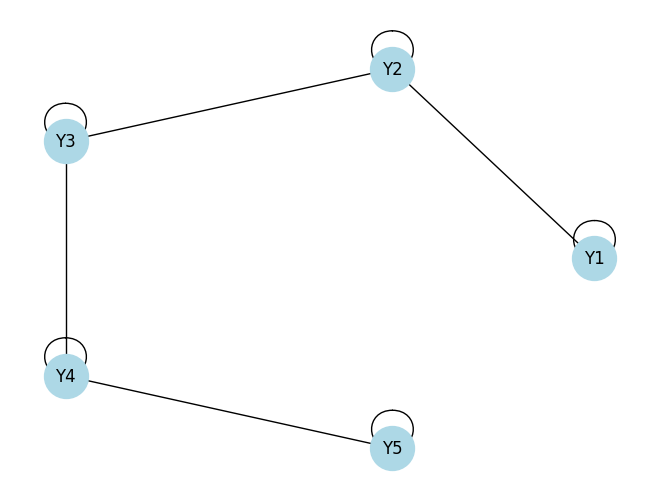}
        \caption{The global graph between each multivariate series $Y_i$: each $Y_i$ interact with $Y_{i-1}$ and $Y_{i+1}$, except $Y_1$ and $Y_5$, following equation \ref{equadiff}.}
    \end{figure}
    \begin{figure}[!h]
        \centering
        \includegraphics[width=0.65\linewidth]{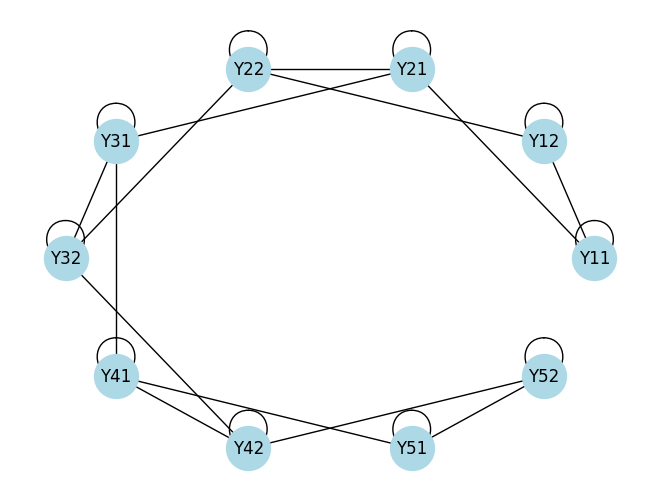}
        \caption{The explicit graph between each univariate time series $Y_{ij}$.}
    \end{figure}
    \begin{figure}[!h]
        \centering
        \includegraphics[width=0.65\linewidth]{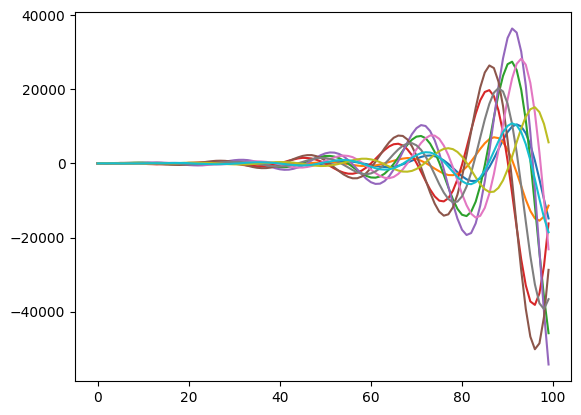}
        \caption{Trajectories $Y_{ij}$ for all $i,j$.}
    \end{figure}
    \begin{figure}[!h]
        \centering
        \includegraphics[width=0.65\linewidth]{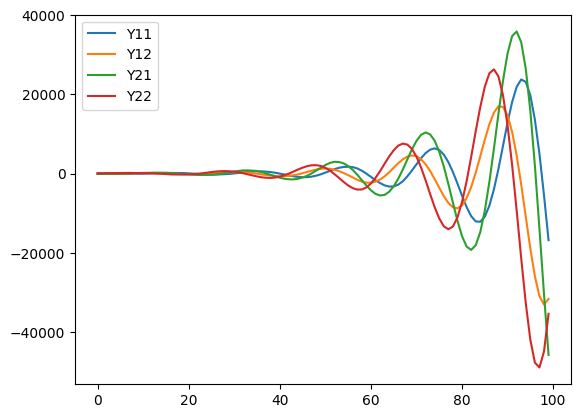}
        \caption{$Y_1$ and $Y_2$ only.}
        \label{fig:enter-label}
    \end{figure}
    
\end{itemize}
As metrics, we compute the averaged proportion of True Positive edges (TPE), True Negative edges (TNE), accuracy, precision, recall and F1-score. We choose to fix hyperparameters (e.g. order of truncated signature) at values that seemed correct with respect to literature (see \cite{fermanian2021learning} Chapter 3). Future work will include a hyperparameters optimization. Our initial experiments yield the results presented in Table \ref{results}. 

\begin{table}[!h]
\centering
\begin{tabular}{|c|c|c|c|c|}
     \hline \textbf{Number of vertices} & $n=5$ & $n=6$ & $n=7$ & $n=8$ \\
     \hline \textbf{TP} & 0.72 & 0.75 & 0.67 & 0.68\\
     \hline \textbf{TN} & 0.62 & 0.68 & 0.68 & 0.72\\
     \hline \textbf{Accuracy} & 0.66 & 0.71 & 0.67 & 0.71\\
     \hline \textbf{Precision} & 0.56 & 0.54 & 0.45 & 0.45\\
     \hline \textbf{Recall} & 0.72 & 0.75 & 0.67 & 0.68\\
     \hline \textbf{F1-score} & 0.63 & 0.63 & 0.54 & 0.54\\
     \hline
\end{tabular}
\caption{Results on different size of vertices set: The results show relatively stable performance across different network sizes, with accuracy remaining around 67-71\%. However, precision decreases notably as network size increases (from 0.56 to 0.45), suggesting challenges in larger networks.}
\label{results}
\end{table}

\section{Conclusion and perspectives.}

Our method extends the scope of the algorithm from \cite{chretien2024leveraging} to collections of multivariate time series. To our knowledge, this is the first method to build high dimensional structure over this type of data. Our preliminary results demonstrate promising performance, with accuracy rates ranging from 66\% to 71\% across different network sizes. 

\begin{itemize}
    \item In future works, metrics must be refined.
    \item We need to optimize hyperparameters: order of truncation for signature, sparsity coefficient for variable selection algorithm, etc.
    \item Only LASSO was used as variable selection algorithm. Other algorithms (e.g. SLOPE \cite{bogdan2015slope}, used in Gaussian graphical model in \cite{riccobello2022sparse}) will be tested in future work.
    \item We need to create synthetic data with hyperedges of dimension 2 ($[v_iv_jv_k]$) in order to test the capacity of our algorithm to predict higher dimensional structure.
\end{itemize}

    \bibliographystyle{splncs04}
    \bibliography{biblio}
\end{document}